\documentclass[runningheads]{llncs}
\usepackage{graphicx}
\usepackage{orcidlink}
\usepackage{todonotes}
\usepackage{booktabs}
\usepackage{subcaption}
\usepackage{multirow}
\usepackage{tcolorbox}

\tcbset{width=\textwidth,boxrule=0pt,colback=yellow,arc=0pt,auto outer arc,left=0pt,right=0pt,boxsep=5pt}

\usepackage{amsmath}

\hypersetup{
    colorlinks,
    linkcolor={red!50!black},
    citecolor={blue!50!black},
    urlcolor={blue!80!black}
}

\usepackage[nolist]{acronym}
\begin{acronym}[ECU]
\acro{mAP}{Mean Average Precision}
\acro{CNN}{Convolutional Neural Network}
\acro{VLAD}{Vector of Locally Aggregated Descriptors}
\acro{WR}{Writer Retrieval}
\acro{GMP}{Generalized Max Pooling}

\end{acronym}

\begin{document}
\title{KaiRacters: Character-level-based Writer Retrieval for Greek Papyri}

%
%
\author{Marco Peer\inst{1}\orcidlink{0000-0001-6843-0830} \and Robert Sablatnig\inst{1}\orcidlink{0000-0003-4195-1593} \and Olga Serbaeva\inst{2}\orcidlink{0000-0002-0103-8284}
 \and Isabelle Marthot-Santaniello\inst{2} \orcidlink{0000−0003−0407−8748}}

\authorrunning{M. Peer et al.}
%
\institute{Computer Vision Lab, TU Wien, Austria \and
Departement Altertumswissenschaften, Universität Basel, Basel, Switzerland \\
\email{mpeer@cvl.tuwien.ac.at}\\
}
\maketitle              
\begin{abstract}
This paper presents a character-based approach for enhancing writer retrieval performance in the context of Greek papyri. Our contribution lies in introducing character-level annotations for frequently used characters, in our case the trigram \textit{kai} and four additional letters ($\varepsilon, \kappa, \mu, \omega$), in Greek texts. We use a state-of-the-art writer retrieval approach based on NetVLAD and compare a character-level-based feature aggregation method against the current default baseline of using small patches located at SIFT keypoint locations for building the page descriptors. We demonstrate that by using only about 15 characters per page, we are able to boost the performance up to 4~\% mAP (a relative improvement of 11~\%) on the GRK-120 dataset. Additionally, our qualitative analysis offers insights into the similarity scores of SIFT patches and specific characters. We publish the dataset with character-level annotations, including a quality label and our binarized images for further research.
\end{abstract}

\section{Introduction}
\ac{WR} describes the task of finding documents penned by the same writer as a given query document. Typical applications include forensics or digital humanities, in particular \ac{WR} for historical documents, which also includes papyrology, the study of ancient texts on papyri \cite{grk-papyri}. Scholars face difficulties in assigning writers due to degradation in papyri and variations in handwriting over time. By tracing handwriting evolutions based on solid evidence, \ac{WR} is a possibility to organize papyrological documentation coherently, that is then used for further tasks, e.g., reconstructing ancient archives, refining the accuracy of writer identification or papyri dating \cite{papyrow}. To aid papyrologists through computerized and automated \ac{WR}, deep-learning-based methods have emerged as a promising technique in the field of papyrology \cite{cilia_wi,christlein_papyri,papyrow,peer_fm,pirrone}.

\begin{figure}[t]
\centering 
\includegraphics[width=0.98\textwidth]{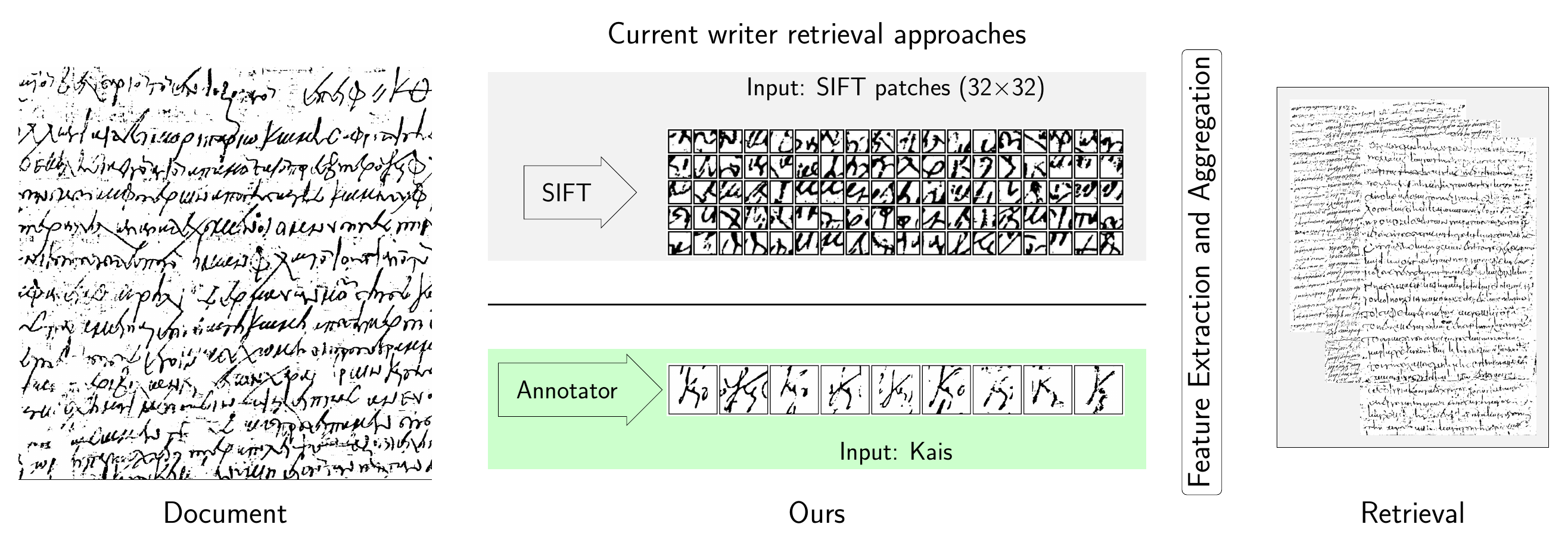}
\caption{Overview of our approach. Contrary to state of the art, we do not aggregate deep features of SIFT patches that usually contain a few strokes of handwriting. Instead, we aggregate features of specific characters, in our case the trigram \textit{kai}, to form the global page descriptor.}\label{fig:kai}
\end{figure}

\ac{WR} methods usually rely on a multi-stage approach: \emph{Sampling} patches of handwriting (e.g. via SIFT keypoint detection \cite{christlein_cnn_vlad,peer_netrvlad}) combined with preprocessing, such as binarization, feature extraction via a neural network and an encoding stage where the statistics of the features are calculated (e.g., \ac{VLAD} \cite{christlein_cnn_vlad} or NetVLAD \cite{peer_netmvlad,peer_netrvlad,rasoulzadeh}) and aggregated. In our paper, we want to extend the current state of the art and investigate the sampling and aggregation step by using character-level annotations of the handwriting to aggregate the NetVLAD-encoded features of specific characters. More specifically, we keep the unsupervised training step, where we cluster SIFT descriptors and use the cluster assignment as a target label of the $32\times 32$ patch extracted at the keypoint, but use only features of specific characters for aggregation during inference for the global page descriptor. An overview of our methodology is given in Fig.~\ref{fig:kai}. While current approaches rely on thousands of local features, extracted by a neural network at SIFT keypoint locations (which we refer to as \emph{SIFT patches}), we investigate the aggregation of the features of specific characters, in our case \emph{kai}, a trigram usually standing for the most frequent word in Greek texts, the conjunction meaning "and". It is also used by scholars to determine authorship of Greek texts \cite{kai1,kai2}. In our work, we concentrate on Greek papyri since this domain still lacks performance for \ac{WR} or writer identification \cite{christlein_papyri,peer_fm}, with \ac{mAP} of only about 40~\%. Additionally to our evaluation, we publish the dataset\footnote{Dataset: \url{https://d-scribes.philhist.unibas.ch/en/case-studies/dioscorus/kairacters/}
} of the full images of the GRK-120 dataset - that was until now only available as segmented rows in \cite{papyrow} along with the annotated characters (bounding boxes and character images in color and binarized) - and we compare performances for other characters (epsilon $\varepsilon$, kappa $\kappa$, mu $\mu$, and omega $\omega$).

We show that by using character-level annotations and aggregating only specific characters, we 1) boost the retrieval performance in terms of \ac{mAP}, even when the quality of the letters is low, and 2) reduce the amount of handwriting needed to obtain a discriminate global page descriptor (11 \textit{kai}-s per page vs. 2600 SIFT patches). Our evaluation is thoroughly conducted on subsets of the GRK-120 dataset, ensuring fair comparisons on the \ac{WR} performance. Furthermore, our qualitative studies show that the use of characters detects complementary similarities of the writers, rather than only improving it.

To summarize, our contributions are:
\begin{itemize}
    \item We release our dataset - the full images as well as the images and annotations of all of our characters used and of the remaining 20 letters - in color and binarized version. For the \textit{kai}-s, quality labels are included.
    \item We evaluate character-level-based feature aggregation with a state-of-the-art \ac{WR} approach \cite{peer_netrvlad} and show that this outperforms the currently dominating methodology of aggregating features of SIFT patches.
    \item Our approach does only need a few samples to achieve similar performances as the baseline using SIFT patches. With only~ 11 \textit{kai}-s, the performance is on par or even better compared to 2.6k SIFT patches per sample.
    \item We qualitatively evaluate the similarity of SIFT patches as well as the \textit{kai-}s, providing insights for scholars and further research.
\end{itemize}

Our paper is structured as follows: Section~\ref{sec:rel_work} describes related work in the field of \ac{WR} for Greek Papyri. In Section~\ref{sec:methodology}, details on the data used and the \ac{WR} approach are given. The evaluation protocol is described in Section~\ref{sec:eval}, and our results are presented in Section~\ref{sec:results}. We conclude our paper in Section~\ref{sec:conclusion}.

\section{Related Work}\label{sec:rel_work}
In the following, we provide a brief overview of previous work on \ac{WR} and writer identification in the domain of papyri.

Pirrone et al. \cite{pirrone} explore a self-supervised method for retrieving papyri fragments utilizing a Siamese network with contrastive loss. Their evaluation includes the Michigan Papyrus Collection and a subset of HisFragIR20. Similar to our work, Christlein et al. \cite{christlein_papyri} implement a previously established algorithm \cite{unsupervised_icdar17}, training a network on clustered SIFT descriptors with the initial, 50 image-large GRK-Papyri dataset\cite{grk-papyri}. They rely on a U-Net-based binarization technique for papyri and demonstrate that eliminating degradation and background artifacts notably enhances the retrieval performance. The GRK-Papyri authors present baseline outcomes utilizing local NBNN \cite{hussein}, a learning-free algorithm based on SIFT descriptors, but struggle with document degradation issues. Nasir et al. \cite{nasir_Wi} concentrate on binarization via Deep Otsu and train a neural network for writer identification using $512\times 512$ patches. In \cite{papyrow}, Cilia et al. introduced PapyRow, an extension of GRK-Papyri from 50 to 120 images released as fragments via line segmentation. In our previous work, we proposed a feature mixing network for the retrieval and identification of fragments \cite{peer_fm}, and show that the learned descriptors still rely on patterns included in the background. Another work of Cilia et al. \cite{cilia_wi} focuses on the writer identification of patches on a slightly extended dataset compared to the initial GRK-papyri (of 50 images from 10 writers to which they added one writer, Dios, and its 15 writing samples from GRK-120) with different convolutional backbones. In 
this work, our \ac{WR} approach relies on a similar binarization technique as Christlein et al. \cite{christlein_papyri} who use a U-Net and a manually designed augmentation pipeline for degradation usually found in papyri. Regarding the deep feature extraction of patches, we use a residual network and NetRVLAD as proposed in our previous work \cite{peer_netrvlad}. 

Greek papyri have been also involved in two recent competitions: the binarization competition DIBCO2019 \cite{dibco2019} and the competition organized by Seuret et al. \cite{Seuret2023} at ICDAR2023, with the latter one primarily focusing on the localization and classification of characters, resulting in promising recognition outcomes. Our paper's contributions build upon those results. Although the characters we used were manually annotated by expert papyrologists, with the advance in character detection and classification in Greek papyri, our method can in the future be applied to further enhance \ac{WR} performance and eliminate the need for manual annotation by scholars. Another recent task on Greek papyri is their chronological attribution, or estimating their date, where, e.g., Pavlopoulos et al. \cite{papyri_dating} apply a method based on \acp{CNN}.

\section{Methodology}\label{sec:methodology}
In this section, we describe the main parts of our methodology. We start with the description of the data, in particular the statistics including the characters, followed by the \ac{WR} pipeline.

\subsection{Data}
The dataset used in this paper is based on the GRK-120 dataset, consisting of 120 documents belonging to 23 different writers that have been identified so far in the course of the D-Scribes\footnote{\url{https://d-scribes.philhist.unibas.ch/en/}} project. One of the aims of this project is the computer-assisted identification of scribes of the archive of Dioscorus of Aphrodito, the richest papyrus archive of the Byzantine period \cite{grk-papyri}. Secondly, the characters of the texts are annotated as two distinct subsets - 1300 \textit{kai-}s and 9511 individual characters representing one of the 24 letters of the Greek alphabet. For our work described below, we only use part of it, but make the full data available for further research. 

\begin{table}[t]
\centering
\caption{Statistics of the annotations for the characters and \textit{kai-}s (BT\texttt{x} describes the quality label). We also show randomly sampled examples of the annotated characters.}\label{tab:letter_statistics}
\begin{tabular}{cccc}
\toprule
\multicolumn{2}{l}{Character} & Samples & Examples\\ \midrule
\multicolumn{2}{l}{$\varepsilon$}  & 504 & \includegraphics[width=0.7\textwidth]{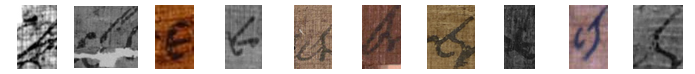} \\
\multicolumn{2}{l}{$\kappa$}   & 353 & \includegraphics[width=0.7\textwidth]{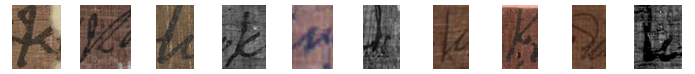} \\
\multicolumn{2}{l}{$\mu$}  & 287 & \includegraphics[width=0.7\textwidth]{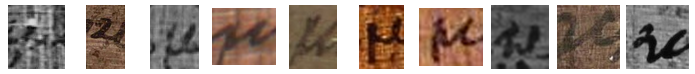}\\
\multicolumn{2}{l}{$\omega$}  & 318 &  \includegraphics[width=0.7\textwidth]{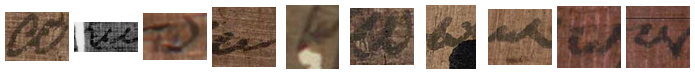}\\ \midrule
~ & {\scriptsize BT1} & 720&  \includegraphics[width=0.7\textwidth]{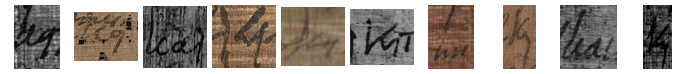}  \\
\textit{kai} &  {\scriptsize BT2} & 380&  \includegraphics[width=0.7\textwidth]{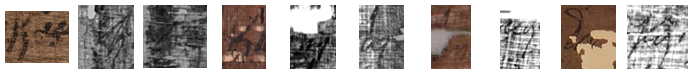} \\
~ &  {\scriptsize BT3} & 51&  \includegraphics[width=0.7\textwidth]{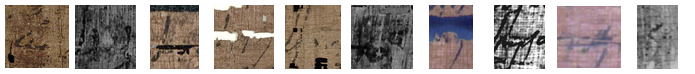}  \\ \bottomrule
\end{tabular}
\end{table}

\paragraph{Annotations}
The annotations of the characters of the texts in GRK-120 are done with the READ software\footnote{\url{https://github.com/readsoftware/read}}. We follow the workflow described in our previous work \cite{serbaeva} and define two different sets of annotations, indicated in Table~\ref{tab:letter_statistics}, which are publicly available: 
\begin{itemize}
    \item The first set consists of letters, usually of the best-preserved three lines of text. Since the annotations are not entirely complete, we provide only preliminary results using four of those characters ($\varepsilon$, $\kappa$, $\mu$, $\omega$). They are chosen by us since they are assumed to be discriminative for the handwriting of the scribes \cite{stylistic_similarity}.
    \item The second set includes only \textit{kai-}s, a trigram formed of kappa $\kappa$, alpha $\alpha$ and iota $\iota$ that in most of the cases stands for the most frequent word in the Greek language (corresponding to the English "and") but occasionally also occurs as part of a word (for instance \textit{dikaios}, "just, fair"). This trigram was chosen not only because of its very high frequency that makes it likely to appear even in small papyri, but also because its shape is usually not affected by the previous or following character (no ligature before $\kappa$ nor after $\iota$). The quality label of each \textit{kai} is additionally provided.
\end{itemize}
 


\paragraph{Quality labeling} The \textit{kai-}s are tagged according to their preservation state. We provide three labels (\texttt{BT\{1,2,3\}}), where \texttt{BT1} indicates best quality (no degradation), annotations tagged with \texttt{BT2} are partly damaged, and unreadable \textit{kai-}s are assigned with \texttt{BT3}, where the \textit{kai} is only identifiable given the context of the text. Examples and the quantity of each label are given in Table~\ref{tab:letter_statistics}.


\begin{figure}[t]
\centering
\includegraphics[width=\textwidth]{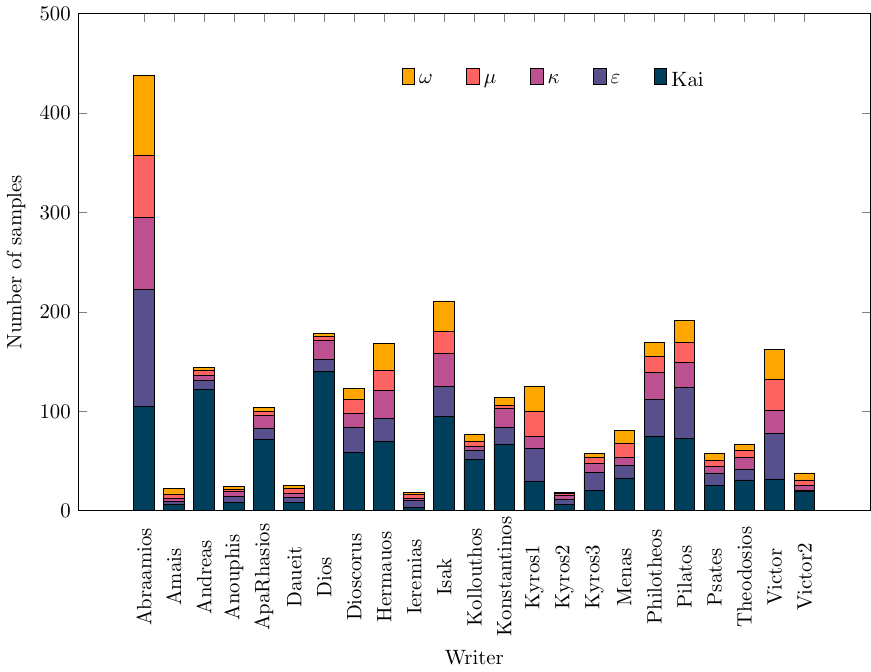}
\caption{Distribution of the characters used per writer.}\label{fig:char_dist}
\end{figure}


\paragraph{Final dataset} The final dataset consists of \textit{kai-}s and the characters $\varepsilon$, $\kappa$, $\mu$ and $\omega$, and those are assigned to a writing sample (an image containing only one writer's handwriting) from the GRK-120 dataset.
Note that both set of annotations
are independent, e.g., the $\kappa$-s used for aggregation might be part of the \textit{kai-}s.
This assignment allows us to compare the traditional \ac{WR} approaches to our proposed character-based aggregation scheme. We provide \textit{kai-}s for each of the 23 writers in the dataset. Since the number of pages the writers contributed in the GRK-120 dataset are highly imbalanced (e.g. Abraamios has 21 writing samples from 18 papyri), we also observe an uneven distribution of characters per writer, which increases the difficulty of \ac{WR} by having only a small amount of handwriting for some writers, e.g. Kyros1 or Victor2. The distribution of the characters is also shown in Fig.~\ref{fig:char_dist}. Due to the presence in the GRK-120 dataset of short and damaged writing samples, and a limited time for manual annotation, we are not able to provide \textit{kai-}s or letters for each document. We provide more details on the subsets we evaluate our approach on in Section~\ref{sec:eval}.

\subsection{Writer Retrieval}
In this section, we describe our method for \ac{WR}. We start with preprocessing, which mainly consists of binarization, followed by the feature extraction and aggregation part.

\begin{figure}
\centering
\begin{subfigure}[b]{\textwidth}
\centering
\includegraphics[width=\textwidth]{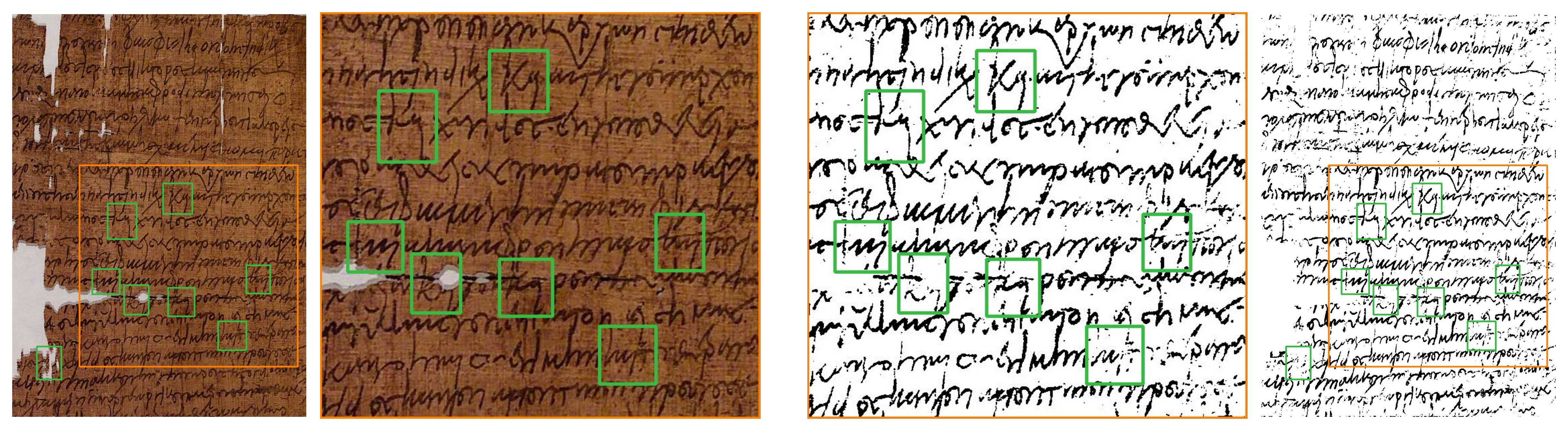}
\end{subfigure}
\begin{subfigure}[b]{0.5\textwidth}\centering
\includegraphics[width=0.98\textwidth]{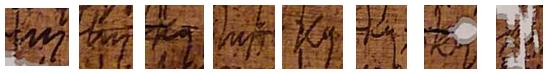}
\end{subfigure}\hfill
\begin{subfigure}[b]{0.5\textwidth}\centering
\includegraphics[width=0.98\textwidth]{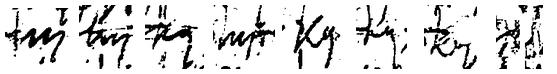}
\end{subfigure}
\caption{Example page with annotated \textit{kai-}s (green) in color and binarized. Note that we use a separate U-Net for binarizing characters and full pages (ID: Isak\_5).}\label{fig:color_binarized}
\end{figure}

\paragraph{Binarization} Given that our method is tailored for binarized handwriting and that Greek papyri typically exhibit severe degradation, such as artifacts, holes, or background patterns, we employ binarization as a preprocessing step to isolate the handwriting. Following the approach of Christlein et al. \cite{christlein_papyri}, we utilize a U-Net-based binarization technique, which has been demonstrated to outperform traditional methods for papyri. To address the  binarization of full pages as well as crops of the annotated characters - parts of the characters might suffer from poor binarization, e.g. as shown in Fig.~\ref{fig:color_binarized} - we apply two training strategies, both trained on grayscale images from the DIBCO2019 dataset \cite{dibco2019}, subset II that contains ten papyri:

\begin{enumerate}
    \item For full pages, the binarized output may contain entirely white regions without any handwriting. We train a U-Net on randomly sampled $128\times 128$ patches.
    \item For images of characters, we assume accurate annotations and, therefore, no empty patches. Consequently, we focus our patch sampling exclusively on regions containing handwriting. Moreover, we train on patch sizes ranging from 32 to 128.
\end{enumerate}

Examples of the two binarization methods, as well as the binarized \textit{kai-}s, are shown in Fig.~\ref{fig:color_binarized}. Both networks are able to segment the handwriting and remove most of the artifacts from the papyri.

\paragraph{Feature Extraction} The retrieval method we use is based on the work proposed in \cite{peer_netrvlad}. It consists of a ResNet20 for feature extraction and NetRVLAD to encode the features. The network is trained in an unsupervised manner by a two-step approach: SIFT keypoints are detected, and the corresponding descriptors are clustered. We then use the cluster assignment as a surrogate label for training, where a $32\times 32$ patch serves as input to the network. 

\paragraph{Aggregation} We calculate the global page descriptor $\boldsymbol{X}$, of a page consisting of $N$ samples of the NetRVLAD-encoded set of features $\mathcal{F} = \{\boldsymbol{x}_0, \boldsymbol{x}_1, \cdots, \boldsymbol{x}_{N-1}\} $ by sum-pooling $\boldsymbol{X} = \sum_{i=0}^{N-1} \boldsymbol{x}_i$ followed by element-wise power normalization $f(x) = \mathrm{sign}(x) |x|^\alpha$ with $\alpha=0.4$. Finally, we whiten the page descriptors and $l_2$-normalize again.

The key idea of our paper is that instead of following state-of-the-art approaches and performing inference using $32\times 32$ patches extracted at SIFT keypoints, we aggregate a feature set $\mathcal{F}$ of previously defined characters for \ac{WR}. Although our work relies on manual annotation, automated approaches, e.g., character classification, are an active field of research \cite{Seuret2023} and could be integrated into our pipeline to eliminate the need for human annotations. 

In contrast to SIFT patches, which only contain a few strokes, we investigate the use of larger parts of handwriting, such as characters or trigrams (\textit{kai-}s), since our network is not restricted to a specific image sizes. Since the annotated characters are usually larger than the strokes in the SIFT patches, we use a slightly higher image size ($64\times 64$) for the characters, compared to size $32\times32$ of the SIFT patches. We also conducted experiments for SIFT patches using higher image sizes, but this did not yield significant differences in performance. 


\section{Evaluation}\label{sec:eval}

\begin{figure}[t]
    \centering
    \includegraphics[width=0.9\textwidth]{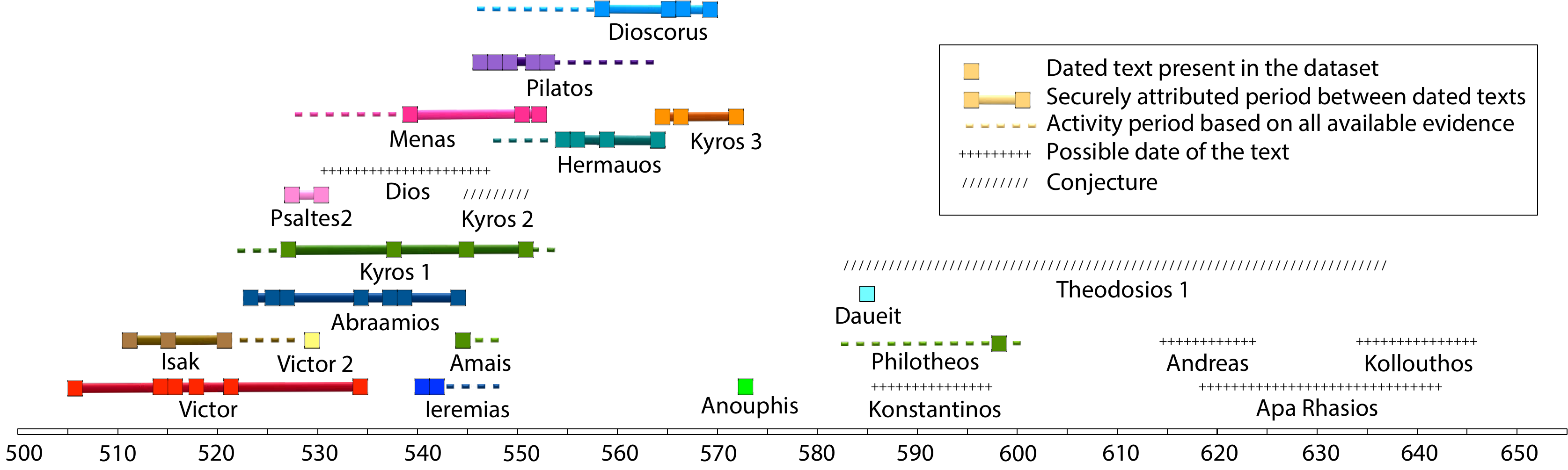}
    \caption{Timeline of the documents included in GRK-120. The time of origin of some documents is still unknown.}
    \label{fig:timeline}
\end{figure}

\paragraph{Dataset}  The GRK-120\footnote{\url{https://d-scribes.philhist.unibas.ch/en/case-studies/dioscorus/}} dataset consists of 120 writing samples penned by 23 writers, with the number of documents per writer being highly unbalanced. Since the documents are contracts, the authorship of these samples is verified by experts' examination of scribes' signatures. Furthermore, the documents vary regarding their date, a timeline is shown in Fig.~\ref{fig:timeline}. The corpus dates from the 6th and 7th century CE, where the date of some papyri, e.g. written by Theodosius or Kollouthos, is still only an estimate.

\begin{table}
\renewcommand{\arraystretch}{1.1}
\caption{Variants of GRK-Papyri dataset used in our evaluation.}
\label{tab:datasets}
\centering
\vspace{5pt}
\begin{tabular}{lcl}
\toprule
Set &  ~ Writers ~ & ~ \\ \midrule
GRK-50 & 10 & Baseline dataset for comparison with state of the art \cite{grk-papyri}\\
GRK-69 & 23 & Evaluation of characters $\varepsilon$, $\mu$, $\kappa$, $\omega$ \\ 
GRK-110 & 23 & Evaluation of \textit{kai-}s  \\
GRK-120 & 23 & Full dataset, extension of GRK-50 \\
\bottomrule
\end{tabular}
\end{table}

We report our results on the GRK-120 dataset and, due to the limited availability of letters on specific pages, subsets of GRK-120, as shown in Table~\ref{tab:datasets}. We use the GRK-69 subset for the evaluation of combinations of letters and the GRK-110 version to evaluate the \textit{kai-}s. \textit{Kai-s} are more numerous because they have been carefully looked for and exhaustively annotated .

\paragraph{Training} We mainly follow the approach in \cite{peer_netrvlad}: We use ResNet20 and append NetRVLAD with 32 clusters as a feature extractor. Our networks are trained in an unsupervised manner by applying Cl-S \cite{unsupervised_icdar17} with 5000 clusters, hence we do not need any writer labels. For training, we use Adam optimizer for 30 epochs with a learning rate of $10^{-4}$ and a batch size of 1024.  

\paragraph{Evaluation Protocol} Our approach is evaluated by using leave-one-image-out cross validation. Each document of the dataset is once used as a query, and the remaining documents are ranked according to the cosine similarity of the page descriptors. Our main metric used is \ac{mAP} that considers the full ranked list. Furthermore, similar to \cite{christlein_papyri}, we report (soft) Top-1, Top-5, Top-10 scores. Top-$x$ indicates if at least one document written by the same writer is included in the first $x$ documents of the ranked list. Finally, we also calculate the precision at $k$ (Pr@$k$), which describes the ratio of correct documents in the first $k$ elements. All of our results are reported on an average of five runs with different seeds.

\section{Results}\label{sec:results}

In this section, we provide our main results. First, we start by comparing our method to state of the art, and follow with a description of results when using only predefined characters for \ac{WR}. Finally, we also show qualitative results of the retrieval.

\subsection{Baseline Performance}

\paragraph{GRK-50} Firstly, to show the effectiveness of our method, we provide results when aggregating SIFT patches (referred to as baseline) on the GRK-50 \cite{grk-papyri} dataset. The state-of-the-art method is proposed by Christlein et al. \cite{christlein_papyri}, using Cl-S training as well as SIFT descriptors, both encoded by mVLAD. As shown in Table~\ref{tab:grk50-baseline}, our approach has a slightly lower mAP, but outperforms \cite{christlein_papyri} regarding Top-1 accuracy. Additionally, we check if training on more data helps our network by training on the GRK-120 dataset. Interestingly, we observe a drop in performance on each metric reported which might be a result of complexity added to the training process, e.g. an increased amount of variety in handwriting by including more writers.

\begin{table}
\renewcommand{\arraystretch}{1.1}
\caption{GRK-50: Comparison to state of the art.}
\label{tab:grk50-baseline}
\centering
\vspace{5pt}
\begin{tabular}{lcccc}
\toprule
Method &  Top-1 & Top-5 & Top-10 & mAP  \\ \midrule
Cl-S + ResNet20 + NetRVLAD & \textbf{58.0} & 74.0 & \textbf{94.0} & 39.1 \\ 
Cl-S + ResNet20 + NetRVLAD {\tiny (trained on GRK-120)} & 52.0 & 74.0 & 90.0 & 38.2 \\ \midrule
R-SIFT + mVLAD \cite{christlein_papyri} & 48.0 & \textbf{84.0} & 92.0 & \textbf{42.8} \\  
Cl-S + ResNet20 + mVLAD \cite{christlein_papyri} & 52.0 & 82.0 & \textbf{94.0} & 42.2 \\  
\bottomrule
\end{tabular}
\end{table}

\paragraph{GRK-120 and subsets} Next, we report the performance of the baseline on the subsets we subsequently use to evaluate our character-based approach. Results are presented in Table~\ref{tab:grk-baseline}. The GRK-69 dataset achieves the lowest mAP value in our experiments (32.4~\%). The decreased number of documents seems to increase the difficulty of the retrieval. The mAP for GRK-110 is slightly better than GRK-120 (39.5~\% vs 39.4~\%), indicating that we remove pages that are hard to retrieve or contain less handwriting - with an increase of 6~\% with respect to Top-1 accuracy. The difference in performance for both subsets when training on the full dataset is less than 1~\% mAP.

\begin{table}
\renewcommand{\arraystretch}{1.1}
\caption{Baseline results on GRK-69, GRK-110 and GRK-120.}
\label{tab:grk-baseline}
\centering
\vspace{5pt}
\begin{tabular}{llcccc}
\toprule
Test set & Train set & Top-1 & Top-5 & Top-10 & mAP  \\ \midrule
\multirow{2}{*}{GRK-69} & GRK-69 & 50.0 & 59.4 & 71.0 & 32.4  \\ 
 & GRK-120 & 48.3 & 60.9 & 71.0 & 32.3 \\ \midrule
 \multirow{2}{*}{GRK-110} & GRK-110 & 62.5 & 79.1 & 81.0 & 39.5 \\ 
 & GRK-120 & 61.5 & 78.2 & 84.5 & 40.0 \\ \midrule
GRK-120 & GRK-120 & 56.5 & 74.2 & 80.0 & 39.4 \\ 
\bottomrule
\end{tabular}
\end{table}

\subsection{Character-based Aggregation}

The method proposed in this paper is character-based aggregation instead of using SIFT patches. We use the same networks for aggregating the features of the respective samples. 

\paragraph{\textit{Kai-}s} Firstly, we evaluate the annotated \textit{kai-}s, provided with a quality label (\texttt{BTx}), and present the retrieval results in Table~\ref{tab:kais_grk110}. We observe a performance gain of +2.4~\% with respect to the main metric, mAP, on the GRK-110 dataset. However, the Top-$x$ accuracies still trail. Therefore, we also check the precision of the first five/ten documents, where the \textit{kai-}s outperform the SIFT approach, which shows that we have more documents retrieved on top of the list. Secondly, considering the amount of data used to calculate the global page descriptor for each document - about 2.6k per page for random SIFT patches vs. 11 \textit{kai-}s - our experiment shows that the actual text influences the performance. By using pre-defined characters as input, in our case \textit{kai-}s, we improve the retrieval while also reducing the amount of data needed. Finally, we find that the retrieval is able to benefit even from \textit{kai-}s of lower quality (e.g. damaged ones), where we achieve the highest mAP for considering all samples (BT\{1,2,3\}). We argue that this might be due to pages containing a small amount of samples, and adding data might be beneficial, even if parts of the character are not fully available.

\begin{table}
\renewcommand{\arraystretch}{1.1}
\caption{Retrieval results on GRK-110 for using \textit{kai-}s or SIFT patches (Baseline).}
\label{tab:kais_grk110}
\centering
\vspace{5pt}
\begin{tabular}{lccccccc}
\toprule
Kais &  Top-1 & Top-5 & Top-10 & Pr@5 & Pr@10 & mAP & Samples per page  \\ \midrule
BT1 & 47.0 & 67.3 & 75.0 & 31.7 & 24.8 & 38.6 & 6.5 \\
BT\{1,2\} & 49.0 & 70.0 & 79.0 & \textbf{36.0} & 29.4 & 41.1 & 11.0\\
BT\{1,2,3\} & 50.0 & 67.3  & 76.4 & 35.8 & \textbf{29.8} & \textbf{41.9} & 11.5 \\ \midrule
Baseline & \textbf{62.5} & \textbf{79.1} & \textbf{81.0} & 35.8 & 26.2 & 39.5 & 2614.8\\ 
\bottomrule
\end{tabular}
\end{table}

\paragraph{Baselines for \textit{Kai}-s} 
Secondly, we evaluate our character-based approach on other baseline methods as well, namely different encodings and aggregations. Since there is no independent training set available, we rely on ResNet56 trained via Cl-S \cite{unsupervised_icdar17} as an unsupervised feature extractor. The results for NetVLAD, SumPooling, \ac{GMP}~\cite{GMP} and \ac{VLAD} are shown in Table~\ref{tab:other_baselines}. They show that the aggregation of \textit{kai}-s perform on par or slightly outperform the SIFT patches in terms of \ac{mAP}, with \ac{VLAD} being the only method where the SIFT patches work better. We assume that this is mainly because the training set used for the vocabulary only consists of the random SIFT patches, with no training set for the \textit{kai}-s available. This also demonstrates the benefit of the integrated codebook learned during training of NetVLAD, which seems to generalize better than the codebook of VLAD, generated by k-means. However, for Greek papyri, advanced encoding strategies such as NetVLAD or VLAD are not significantly better than just SumPooling. Furthermore, the results show that \textit{kai}-s are able to eliminate the need for data, with only 11 \textit{kai}-s performing similarly to 2.6k SIFT patches per page.

\begin{table}
\renewcommand{\arraystretch}{1.1}
\caption{
Other baseline results on GRK-110 for using \textit{kai-}s or SIFT patches.}
\label{tab:other_baselines}
\centering
\vspace{5pt}
\begin{tabular}{lcccc}
\toprule
~ & \multicolumn{2}{c}{Baseline} & \multicolumn{2}{c}{Kais} \\  \cline{2-3} \cline{4-5}
~ &  Top-1 & mAP  & Top-1 & mAP \\ \midrule
NetVLAD & \textbf{62.5} & 39.5 & 50.0 & 41.9 \\
SumPooling & 56.7 & \textbf{42.9} & \textbf{61.5} & \textbf{43.0}\\
\ac{GMP} & 56.7 & 42.8 & \textbf{61.5} & 42.8\\ 
\ac{VLAD} & 56.8 & 42.1 & 39.4 & 31.0 \\
\bottomrule
\end{tabular}

\end{table}

\paragraph{Kai-s and other characters} Additionally to the study of the \textit{kai-}s, we provide preliminary results for combining the \textit{kai-}s with the letters mentioned in the methodology: $\varepsilon, \kappa, \mu, \omega$. The dataset used is the GRK-69 subset. We experiment with aggregating all possible permutations of the five characters and report single character performances as well as the best/worst five combinations of the characters in Table~\ref{tab:agg_grk69}. Firstly, as shown in Table~\ref{tab:single_chars}, the use of the \textit{kai-}s also outperforms the baseline of using SIFT patches by 3.2~\%. Other characters perform worse, which we assign to the lack of data or the fact that they are easily distorted by the previous or following character (due to the cursivity of the handwriting). However, evaluating combinations of different letters, we find that we are able to boost the retrieval performance, with the combination of \textit{kai-}s and $\kappa$ performing the best (36.2~\% mAP, Table~\ref{tab:multi_chars_best}). For a fair overview, we also give the five worst combinations, for which none of them includes the \textit{kai-}s. We conclude that the \textit{kai-}s contain the most discriminating power for \ac{WR}, but annotating or automatically detecting specific characters for aggregation improves the performance, even if we have less data available, e.g., our dataset only includes 287 samples of $\mu$, averaging only four samples per document.

\begin{table}[h]
\caption{Retrieval results on GRK-69 for using different characters or SIFT patches (Baseline). (\subref{tab:single_chars}) Performance when only using one type of character. (\subref{tab:multi_chars_best}) Combinations with best performance. (\subref{tab:multi_chars_worst}) Combinations with worst performance.} \label{tab:agg_grk69}
\centering
\hfill
\begin{subtable}[h]{0.3\textwidth}
\caption{}\label{tab:single_chars}
\centering
\begin{tabular}{lc}
\toprule
Character   & mAP \\ \midrule
Kai & \textbf{35.6} \\
$\varepsilon$ & 23.6 \\
$\mu$ & 24.9 \\
$\kappa$ & 25.1 \\
$\omega$ & 22.3 \\ \midrule
Baseline & 32.4\\ 
\bottomrule
\end{tabular}
\end{subtable}
\hfill
\begin{subtable}[h]{0.3\textwidth}\caption{}\label{tab:multi_chars_best}\centering
\begin{tabular}{lc}
\toprule
Characters   & mAP \\ \midrule
Kai, $\kappa$ & \textbf{36.2} \\
Kai, $\kappa$, $\mu$ & 35.9\\
Kai, $\mu$ & 35.2\\
Kai, $\omega$ & 34.1 \\
Kai,  $\varepsilon$, $\kappa$, $\mu$ & 32.9 \\ \midrule
Baseline & 32.4\\ 
\bottomrule
\end{tabular}
\end{subtable}
\hfill
\begin{subtable}[h]{0.3\textwidth}
\centering
\caption{}\label{tab:multi_chars_worst}
\begin{tabular}{lc}
\toprule
Characters   & mAP \\ \midrule
$\varepsilon$, $\kappa$, $\mu$ & 24.0 \\
 $\varepsilon$, $\kappa$, $\omega$ & 25.1 \\ 
 $\varepsilon$, $\mu$ & 25.3 \\
  $\varepsilon$, $\omega$ & 25.7 \\
  $\kappa$, $\omega$ & 26.3 \\
\midrule
Baseline & \textbf{32.4}\\ 
\bottomrule
\end{tabular}
\end{subtable}
\hfill
\end{table}

\subsection{Qualitative Analysis}
\begin{figure}[t]
\centering
\begin{subfigure}[b]{0.5\textwidth}\centering
\includegraphics[width=0.98\textwidth]{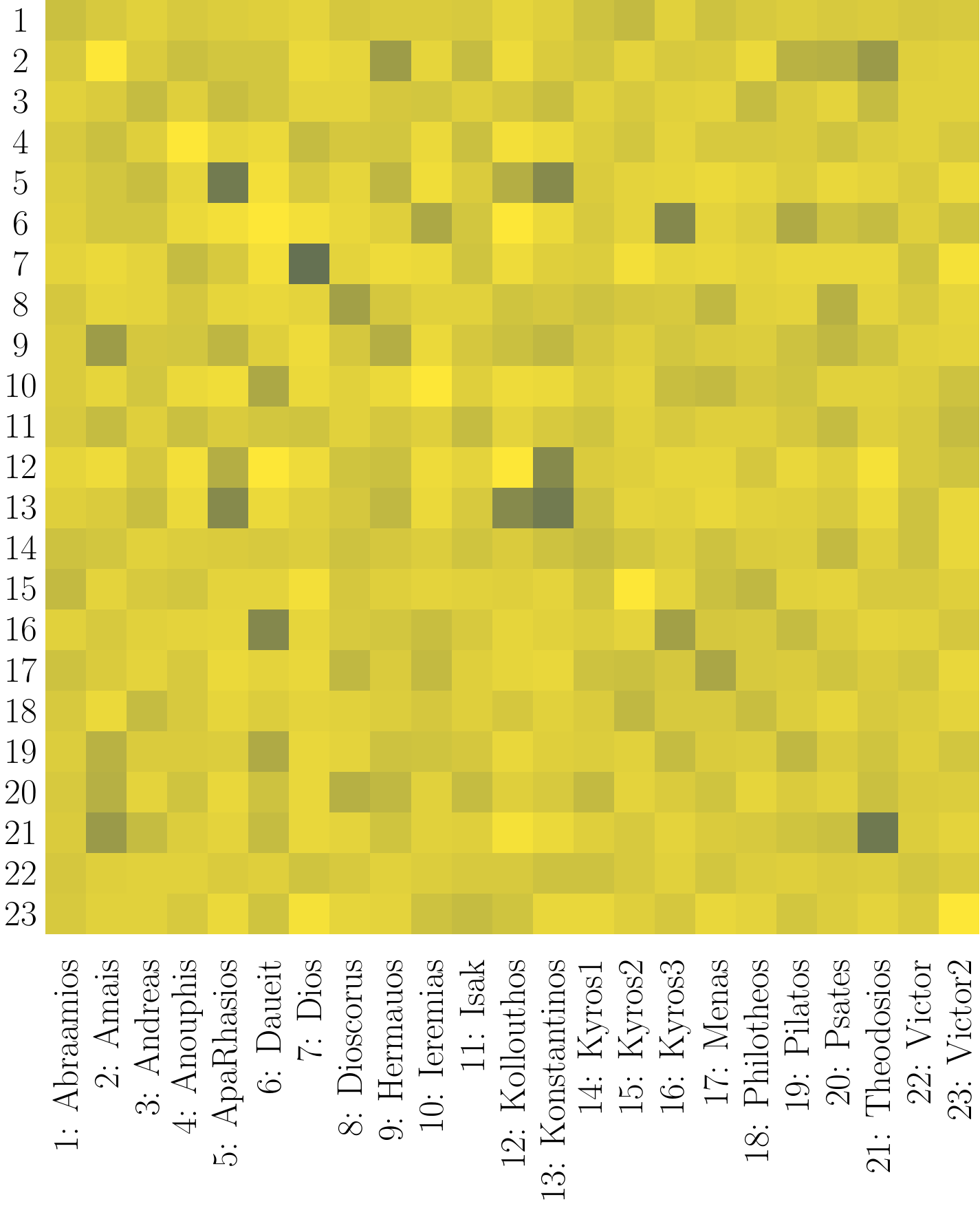}
\caption{SIFT patches}\label{fig:sim_sift}
\end{subfigure}\hfill
\begin{subfigure}[b]{0.5\textwidth}\centering
\includegraphics[width=0.98\textwidth]{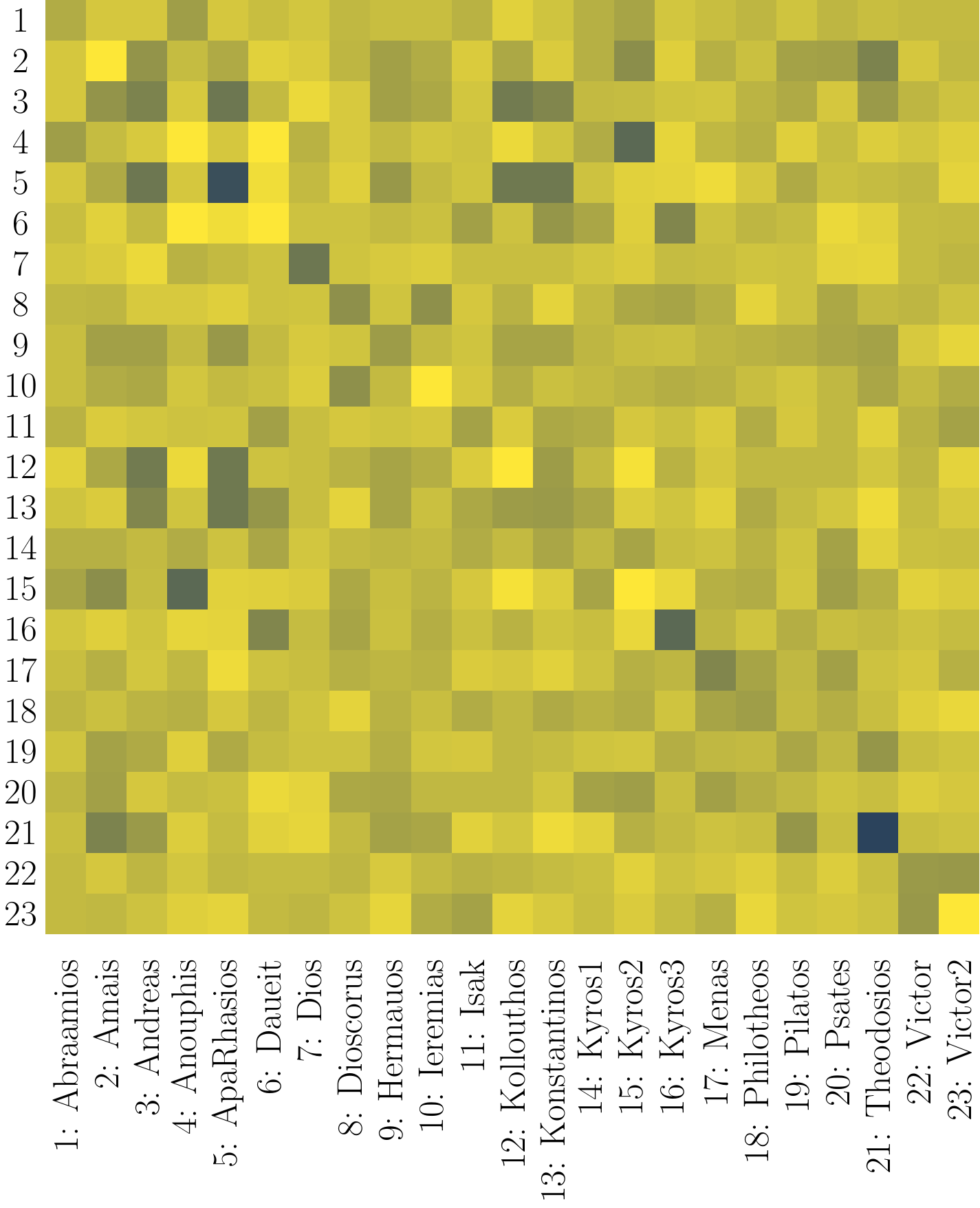}
\caption{Kais}\label{fig:sim_kai}
\end{subfigure}
\caption{Similarity matrices per writer (pairwise document similarity averaged) when using (\subref{fig:sim_sift}) SIFT patches and (\subref{fig:sim_kai}) only \textit{kai-}s for calculating the global page descriptor. The darker the square, the higher the similarity.}\label{fig:qual1}
\end{figure}

We conclude the evaluation of our paper by providing two qualitative studies, both concerning the performance of the \textit{kai-}s. Firstly, we qualitatively compare the writer similarity for SIFT patches and the \textit{kai-}s. We average the pairwise similarity of all documents (excluding pairs for the same documents), and provide a heatmap in Fig.~\ref{fig:qual1}. By comparing the matrices, we observe that some writers have high similarities in both, e.g. Theodosius, where in particular the \textit{kai-}s are highlighted, or Apa Rhasios. However, the intra-class similarity of the \textit{kai-}s is higher for writers like Andreas or Victor than aggregating SIFT patches. This observation is of interest for specialists of ancient handwritings (paleographers) because it illustrates the potential of our approach to contribute characterizing intra- and inter- writer variations in human understandable terms. In general, we also observe multiple instances of similarities across writers in both matrices, which invites papyrologists to examine if indeed those writers have similar styles. It might also be due to a lack of data for the \textit{kai-}s and the general performance in terms of \ac{mAP}, that is only about 40~\%. We find that the aggregation of \textit{kai-}s leads to additional similarities in the feature space for some writers rather than just higher similarities compared to the SIFT patches, indicating that an extension of our approach might consider both aggregation schemes for better performance.

\begin{figure}[t]
\centering
\includegraphics[width=0.55\textwidth]{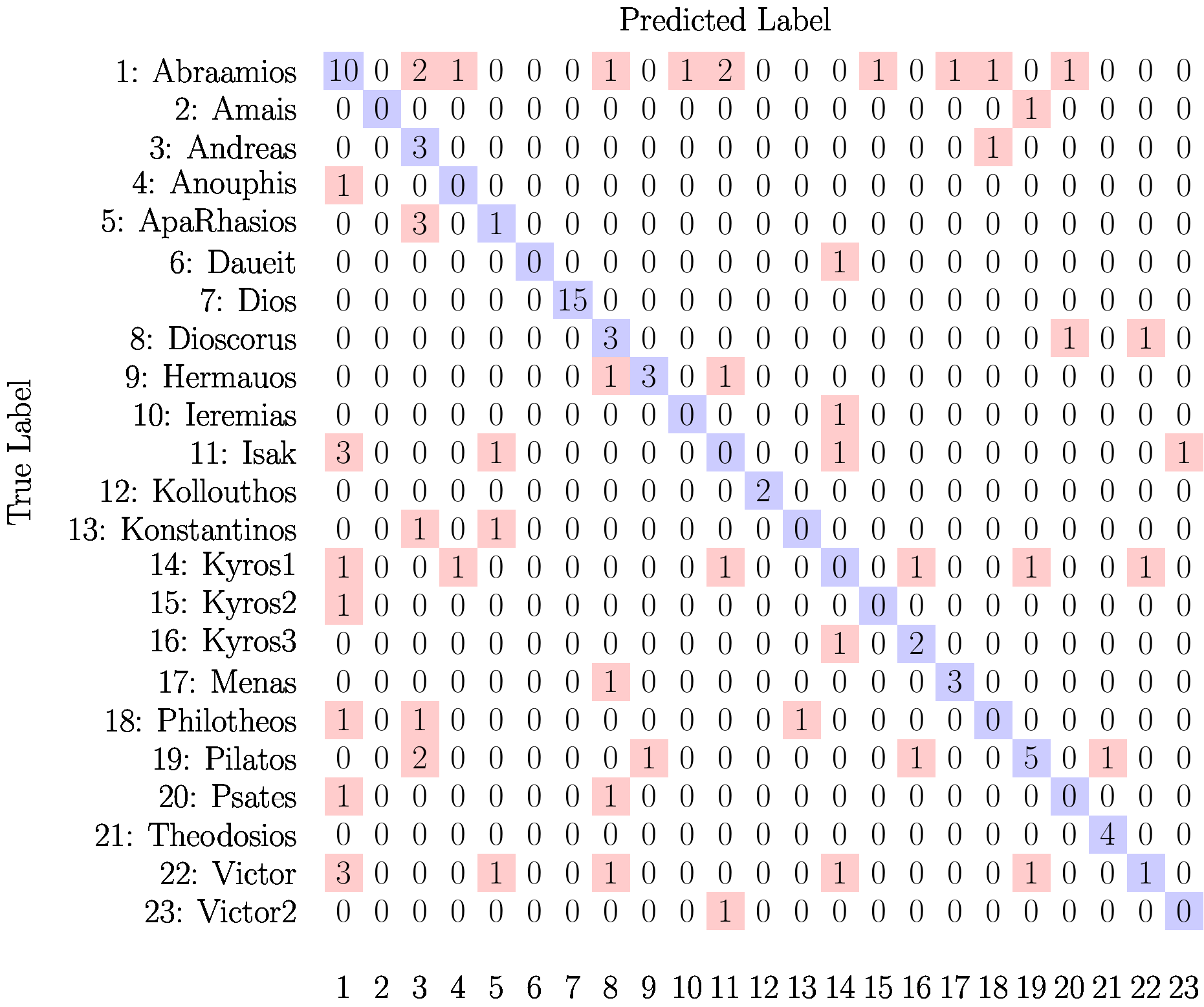}
\caption{Confusion matrix of the writers for each document considering Top-1 accuracy when aggregating the \textit{kai} features.}\label{fig:conf_matrix}
\end{figure}

Secondly, we provide a confusion matrix of the \ac{WR} performance of the \textit{kai-}s where the "predicted label" is the writer of the most similar writing sample, as shown in Fig.~\ref{fig:conf_matrix}. With this, we can identify high similarities between pairs of writing samples. In GRK-120, 10 writers have only one document, but some are long enough to be split into several images or writing samples (from 2 to 15). Interestingly, the writers who have the highest scores, e.g. all their writing samples correctly attributed, are writers attested in one single document of large size (Dios and Theodosios with respectively 15 and 4 writing samples). The five writers that are attested in one single but short papyrus, Amais, Anouphis, Daueit, Kyros2, Victor2, are not correctly recognized (see also Fig.~\ref{fig:char_dist} for the amount of annotations per writer). At the opposite, some writers attested on many documents over a long period of time, like Abraamios and Victor1, are not well recognized either. Thus the method seems to work very well when there is enough data to grasp the original features of one document but does not manage to recognize correctly a hand that probably changes over time (and possibly writing implements). We also note that while some confusions can be explained by chronology such as Apa Rhasios being incorrectly classified three times as Andreas, a coeval notary, some others are with much older or later writers. For instance Dioscorus, who is supposed to have an easy-to-distinguish handwriting, is correctly recognized 3 times but confused once with Victor1 and another time with Psates, who are much earlier. It could be a way for paleographers to spot writers using "archaic" or "innovative" styles. 

\section{Conclusion} \label{sec:conclusion}
This study tackles the challenges of identifying writers in Greek papyri, for which we introduced a new method that focuses on individual characters and found it improves the \ac{WR} performance while reducing also the need of data. We find that the \textit{kai-}s are the most discriminative resource used for feature aggregation, but the \ac{WR} can further be boosted by using other letters ($\kappa$ or $\mu$). Our dataset is publicly available, and we also provided qualitative studies on our approach compared to SIFT patches, the current methodology widely used for \ac{WR}. 
One drawback of our method, the need for a scholar to annotate characters necessary for aggregation, could be overcome by further investigating character localization and classification, which is already an active field of research \cite{Seuret2023}. We think combining our pipeline with a detector is the next step to aggregate all characters of a document for aggregation. It is also worth investigating the aggregation of other $n$-grams, not just \textit{kai}-s. Additionally, this paper only considered simple aggregation techniques, but for future work, our paper opens the field of advanced schemes for aggregating the features of the characters, for example, using graph- or cluster-based methods. In the field of papyrology, \ac{WR}, in particular our character-based approach, is a promising tool to find similarities between documents that have lost information about their context of production (writers but also schools of writers, places and periods), thus allowing major improvements in our understanding of Greco-Roman Egypt.

%
%
%
%
\bibliographystyle{splncs04}
\bibliography{bibliography}

\begin{thebibliography}{10}
\providecommand{\url}[1]{\texttt{#1}}
\providecommand{\urlprefix}{URL }
\providecommand{\doi}[1]{https://doi.org/#1}

\bibitem{unsupervised_icdar17}
Christlein, V., Gropp, M., Fiel, S., Maier, A.K.: Unsupervised feature learning for writer identification and writer retrieval. In: 14th {IAPR} International Conference on Document Analysis and Recognition, {ICDAR} 2017, Kyoto, Japan, November 9-15, 2017. pp. 991--997 (2017)

\bibitem{christlein_cnn_vlad}
Christlein, V., Maier, A.K.: Encoding {CNN} activations for writer recognition. In: 13th {IAPR} International Workshop on Document Analysis Systems, {DAS} 2018, Vienna, Austria, April 24-27, 2018. pp. 169--174 (2018)

\bibitem{christlein_papyri}
Christlein, V., Marthot{-}Santaniello, I., Mayr, M., Nicolaou, A., Seuret, M.: Writer retrieval and writer identification in greek papyri. In: Intertwining Graphonomics with Human Movements - 20th International Conference of the International Graphonomics Society, {IGS} 2021, Las Palmas de Gran Canaria, Spain, June 7-9, 2022, Proceedings. vol. 13424, pp. 76--89 (2022)

\bibitem{cilia_wi}
Cilia, N.D., D'Alessandro, T., Stefano, C.D., Fontanella, F., Marthot{-}Santaniello, I., Molinara, M., di~Freca, A.S.: A novel writer identification approach for greek papyri images. In: Image Analysis and Processing - {ICIAP} 2023 Workshops - Udine, Italy, September 11-15, 2023, Proceedings, Part {II}. pp. 422--436 (2023)

\bibitem{papyrow}
Cilia, N.D., Stefano, C.D., Fontanella, F., Marthot{-}Santaniello, I., di~Freca, A.S.: Papyrow: {A} dataset of row images from ancient greek papyri for writers identification. In: Pattern Recognition. {ICPR} International Workshops and Challenges - Virtual Event, January 10-15, 2021, Proceedings, Part {VII}. vol. 12667, pp. 223--234 (2020)

\bibitem{stylistic_similarity}
Marthot{-}Santaniello, I., Vu, M., Serbaeva, O., Beurton{-}Aimar, M.: Stylistic similarities in greek papyri based on letter shapes: {A} deep learning approach. In: Document Analysis and Recognition - {ICDAR} 2023 Workshops - San Jos{\'{e}}, CA, USA, August 24-26, 2023, Proceedings, Part {I}. vol. 14193, pp. 307--323 (2023)

\bibitem{kai1}
McArthur, H.K.: Kai frequency in greek letters. New Testament Studies  \textbf{15}(3),  339–349 (1969)

\bibitem{hussein}
Mohammed, H.A., M{\"{a}}rgner, V., Konidaris, T., Stiehl, H.S.: Normalised local na{\"{\i}}ve bayes nearest-neighbour classifier for offline writer identification. In: 14th {IAPR} International Conference on Document Analysis and Recognition, {ICDAR} 2017, Kyoto, Japan, November 9-15, 2017. pp. 1013--1018 (2017)

\bibitem{grk-papyri}
Mohammed, H.A., Marthot{-}Santaniello, I., M{\"{a}}rgner, V.: Grk-papyri: {A} dataset of greek handwriting on papyri for the task of writer identification. In: 2019 International Conference on Document Analysis and Recognition, {ICDAR} 2019, Sydney, Australia, September 20-25, 2019. pp. 726--731 (2019)

\bibitem{kai2}
Morton, A.Q.: The authorship of greek prose. Journal of the Royal Statistical Society. Series A (General)  \textbf{128}(2),  169--233 (1965)

\bibitem{GMP}
Murray, N., Perronnin, F.: Generalized max pooling. In: 2014 {IEEE} Conference on Computer Vision and Pattern Recognition, {CVPR} 2014, Columbus, OH, USA, June 23-28, 2014. pp. 2473--2480 (2014)

\bibitem{nasir_Wi}
Nasir, S., Siddiqi, I., Moetesum, M.: Writer characterization from handwriting on papyri using multi-step feature learning. In: Document Analysis and Recognition, {ICDAR} 2021 Workshops, Lausanne, Switzerland, September 5-10, 2021, Proceedings, Part {I}. vol. 12916, pp. 451--465 (2021)

\bibitem{papyri_dating}
Pavlopoulos, J., Konstantinidou, M., Vardakas, G., Marthot{-}Santaniello, I., Perdiki, E., Koutsianos, D., Likas, A., Essler, H.: Explaining the chronological attribution of greek papyri images. In: Discovery Science - 26th International Conference, {DS} 2023, Porto, Portugal, October 9-11, 2023, Proceedings (2023)

\bibitem{peer_netmvlad}
Peer, M., Kleber, F., Sablatnig, R.: Writer retrieval using compact convolutional transformers and netmvlad. In: 26th International Conference on Pattern Recognition, {ICPR} 2022, Montreal, QC, Canada, August 21-25, 2022. pp. 1571--1578 (2022)

\bibitem{peer_netrvlad}
Peer, M., Kleber, F., Sablatnig, R.: Towards writer retrieval for historical datasets. In: Document Analysis and Recognition - {ICDAR} 2023 - 17th International Conference, San Jos{\'{e}}, CA, USA, August 21-26, 2023, Proceedings, Part {I}. pp. 411--427 (2023)

\bibitem{peer_fm}
Peer, M., Sablatnig, R.: Feature mixing for writer retrieval and identification on papyri fragments. In: Proceedings of the 7th International Workshop on Historical Document Imaging and Processing (2023)

\bibitem{pirrone}
Pirrone, A., Beurton{-}Aimar, M., Journet, N.: Self-supervised deep metric learning for ancient papyrus fragments retrieval. Int. J. Document Anal. Recognit.  \textbf{24}(3),  219--234 (2021)

\bibitem{dibco2019}
Pratikakis, I., Zagoris, K., Karagiannis, X., Tsochatzidis, L., Mondal, T., Marthot-Santaniello, I.: Icdar 2019 competition on document image binarization (dibco 2019). In: 2019 International Conference on Document Analysis and Recognition (ICDAR). pp. 1547--1556 (2019)

\bibitem{rasoulzadeh}
Rasoulzadeh, S., BabaAli, B.: Writer identification and writer retrieval based on netvlad with re-ranking. {IET} Biom.  \textbf{11}(1),  10--22 (2022)

\bibitem{serbaeva}
Serbaeva, O., White, S.: Read for solving manuscript riddles: a preliminary study of the manuscripts of the 3rd \emph{\d{s}a\d{t}ka} of the \emph{Jayadrathayāmala}. In: Document Analysis and Recognition – ICDAR 2021 Workshops, Lausanne, Switzerland, September 5–10, 2021 Proceedings, Part 2. pp. 339--348 (2021)

\bibitem{Seuret2023}
Seuret, M., Marthot-Santaniello, I., White, S.A., Serbaeva~Saraogi, O., Agolli, S., Carrière, G., Rodriguez-Salas, D., Christlein, V.: ICDAR 2023 Competition on Detection and Recognition of Greek Letters on Papyri, p. 498–507 (2023)

\end{thebibliography}
\end{document}